# InVDriver: Intra-Instance Aware Vectorized Query-Based Autonomous Driving Transformer

Bo Zhang [a,b], Heye Huang [c], Chunyang Liu [b], Yaqin Zhang [a,d], Zhenhua Xu [a,*]

[a] *School of Vehicle and Mobility, Tsinghua University, Beijing, 100084, China*

[b] *DiDi Global, Beijing, 100081, China*

[c] *University of Wisconsin-Madison, Madison, 53707, USA*

[d] *Institute for AI Industry Research (AIR), Tsinghua University, Beijing, 100084, China*

\* Corresponding author.

E-mail address: zhenhuaxu@tsinghua.edu.cn

**Abstract**

End-to-end autonomous driving with its holistic optimization capabilities, has gained increasing traction in academia and industry. Vectorized representations, which preserve instance-level topological information while reducing computational overhead, have emerged as a promising paradigm. While existing vectorized query-based frameworks often overlook the inherent spatial correlations among intra-instance points, resulting in geometrically inconsistent outputs (e.g., fragmented HD map elements or oscillatory trajectories). To address these limitations, we propose InVDriver, a novel vectorized query-based system that systematically models intra-instance spatial dependencies through masked self-attention layers, thereby enhancing planning accuracy and trajectory smoothness. Across all core modules—perception, prediction, and planning—InVDriver incorporates masked self-attention mechanisms that restrict attention to intra-instance point interactions, enabling coordinated refinement of structural elements while suppressing irrelevant inter-instance noise. Experimental results on the nuScenes benchmark demonstrate that InVDriver achieves state-of-the-art performance, surpassing prior methods in both accuracy and safety, while maintaining high computational efficiency. Our work validates that explicit modeling of intra-instance geometric coherence is critical for advancing vectorized autonomous driving systems, bridging the gap between theoretical advantages of end-to-end frameworks and practical deployment requirements.

## 1. Introduction

Autonomous driving systems represent a sophisticated technological framework comprising integrated subsystems of perception, prediction, motion planning, and vehicular control. This cyber-physical architecture requires the vehicle to continuously interpret multimodal sensory inputs, dynamically model agent behaviors, and execute collision-free trajectories in complex traffic scenarios - demanding not only multi-spectral perception capabilities through sensor fusion but also cognitive-level decision-making algorithms powered by artificial intelligence architectures.

Traditional architectural frameworks for autonomous driving systems employ a sequential modular paradigm comprising discrete functional components (perception, prediction, planning, and control) (Liu et al., 2021; Zablocki et al., 2022; Kendall et al., 2019; Chen et al., 2024; Li et al., 2024). These modules introduce cascading dependencies, where errors from upstream processes propagate, compounding inaccuracies in downstream modules. For instance, the motion planning module necessitates object detection outputs (bounding boxes, semantic segmentation masks) generated by perception modules. While this compartmentalized approach provides initial design transparency and modular development benefits, it introduces three fundamental limitations: (1) Suboptimal system integration. The decoupled optimization strategy permits individual module tuning but fails to ensure global system optimality. The combinatorial interaction of locally optimized components may create emergent error modes unobservable at module level evaluations. (2) Information degradation cascade. Subsequent decision-making layers operate on pre-processed abstractions rather than raw sensor data (e.g., cameras and LiDARs). Critical environmental metadata, including meteorological conditions (precipitation intensity, road friction coefficients) and sensor-specific artifacts (occlusion patterns, multi-path reflections), are systematically discarded during feature extraction, potentially compromising safety-critical decisions. (3) Error propagation amplification. The serialized data pipeline establishes single-point-of-failure vulnerabilities. Perception inaccuracies propagate geometrically through prediction confidence intervals and planning trajectory optimizations.

To address the aforementioned challenges, recent research has shifted toward end-to-end autonomous driving frameworks that optimize the entire system holistically (Xu et al., 2024; Jiang et al., 2023; Hu et al., 2023). Some approaches directly generate planning outputs from raw sensor inputs without intermediate modules (Xu et al., 2024; Pan et al., 2024; Xiao et al., 2020; Zheng et al., 2024; Zeng et al., 2019; Sima et al., 2024; Shao et al., 2024). However, these methods often suffer from inadequate supervision signals, leading to suboptimal performance in complex scenarios. Some other solutions retain modular structures but interconnect them through learnable tensors (Jiang et al., 2023; Hu et al., 2022; Berta et al., 2024; Hu et al., 2023). Unlike conventional systems that produce non-differentiable intermediate representations, these methods establish differentiable connections between modules (e.g., perception to prediction via latent feature tensors), enabling end-to-end gradient propagation and joint optimization of the complete pipeline. Contemporary end-to-end autonomous driving architectures predominantly adopt vectorized representations over rasterized alternatives, primarily motivated by their computational efficiency advantages and enhanced capacity for preserving instance-level semantic features.

While vectorized end-to-end autonomous driving methods demonstrate impressive results, they typically assume points to be independent and identically distributed (i.i.d.). This assumption holds for object detection tasks (e.g., vehicle and pedestrian detection), where distinct objects can reasonably be considered statistically independent (Xu et al., 2024; Liao et al., 2023). However, when applied to structured elements such as lane lines, trajectories, and map features, this premise becomes problematic. For instance, a road lane line composed of 20 sequentially connected points cannot satisfy the i.i.d. condition due to strong spatial correlations between adjacent points. Such intra-instance dependencies are largely overlooked in existing approaches, leading to suboptimal performance. While recent studies have begun exploring intra-instance information utilization in vectorized map detection tasks, no prior work has systematically investigated the role of these structural relationships in end-to-end autonomous driving frameworks.

In this paper, we propose the Intra-instance Vectorized Driving Transformer (InVDriver), a vectorized end-to-end autonomous driving system designed to leverage intra-instance structural relationships across perception, prediction, and planning modules. First, the system employs vector representations for all outputs: (1) Perception module. Map vectors and objects; (2) Prediction module. Predicted trajectories of detected objects; (3) Planning module. Predicted trajectories of the ego vehicle. Then, each vector is represented by an instance query (i.e., preserving object identity) and associated point queries (i.e., capturing spatial features). Finally, all queries are sent to the transformer for end-to-end autonomous driving. Queries from different modules exchange information based on cross-attention layers. Evaluated on the nuScenes benchmark, InVDriver demonstrates superior performance in handling structured road elements compared to existing methods, validating the importance of intra-instance modeling. Our principal contributions are threefold:

(1) We propose a query-based framework that integrates perception, prediction, and planning through learnable vectorized representations, enabling holistic optimization and seamless information exchange.

(2) By leveraging masked self-attention mechanisms, we model intra-instance spatial dependencies in structured elements like lanes and trajectories, improving geometric consistency and motion stability.

(3) Extensive experiments on nuScenes demonstrate that InVDriver surpasses conventional methods in accuracy, safety, and efficiency, highlighting the critical role of intra-instance modeling in vectorized autonomous driving.

## 2. Related Works

### 2.1. End-to-end autonomous driving

Autonomous driving techniques have garnered increasing attention from both academic and industrial communities (Zhang et al., 2024; Singh et al., 2024; Zheng et al., 2024; Chen et al., 2024). In contrast to conventional approaches (Liu et al., 2021; Zablocki et al., 2022; Kendall et al., 2019; Chen et al., 2024; Li et al., 2024; Berta et al., 2024; Li et al., 2024; Wang et al., 2024), end-to-end autonomous driving systems (Jiang et al., 2023; Hu et al., 2022; Berta et al., 2024; Hu et al., 2023; Xu et al., 2024; Pan et al., 2024; Xiao et al., 2020; Zheng et al., 2024; Zeng et al., 2019; Sima et al., 2024; Shao et al., 2024; Cao et al., 2024) demonstrate enhanced integration and global optimization capabilities, featuring streamlined architectures and superior operational performance. The evident advantages of end-to-end systems have spurred numerous recent investigations into novel frameworks and methodologies to maximize driving performance. Current research diverges into two primary directions: Certain approaches directly generate final trajectory plans from raw sensor inputs (Xu et al., 2024; Pan et al., 2024; Xiao et al., 2020; Zheng et al., 2024; Zeng et al., 2019; Sima et al., 2024; Shao et al., 2024), though these methods currently exhibit relatively limited accuracy due to insufficient supervisory signals. Alternative strategies preserve modular system designs while implementing learnable interconnections between components (Jiang et al., 2023; Hu et al., 2022; Berta et al., 2024; Hu et al., 2023), enabling full-system optimization through end-to-end training. Among modular implementations, Jiang et al. (2023) introduced UniAD, widely regarded as the foundational work in modular end-to-end autonomous driving systems. This framework achieves state-of-the-art performance on the NuScenes benchmark through learnable query-based inter-module communication. Building upon this foundation, Jiang et al. (2023) subsequently developed VAD, an enhanced system employing vectorized representations for cross-module information exchange, which has demonstrated superior performance across multiple public autonomous driving benchmarks.

**2.2. Vectorized design for autonomous driving**

The bird's-eye-view (BEV) representation has become a prevalent paradigm in contemporary autonomous driving systems (Liu et al., 2023; Li et al., 2022; Xu et al., 2023). Conventional approaches predominantly employ rasterized data representations for environmental modeling (Li et al., 2022), which encode scene information through dense matrices. While rasterized formats offer straightforward implementation, they impose substantial computational overhead and crucially lack instance-level semantic information - a critical limitation that impedes environmental comprehension by downstream planning modules. To address these challenges, emerging methodologies have adopted vectorized representations (Xu et al., 2021, 2022, 2023, 2024; Liao et al., 2023). These approaches directly generate vector-formatted outputs to model road elements like lanes and boundaries, preserving essential topological relationships while demonstrating enhanced perception capabilities in empirical evaluations. The vector paradigm's structural coherence particularly benefits motion planning algorithms through improved geometric interpretability. Liao et al. (2023) established a significant milestone with MapTR, the first query-based vectorized high-definition (HD) mapping framework. While achieving satisfactory baseline performance, MapTR's assumption of independent and identically distributed (i.i.d.) queries overlooks critical intra-instance correlations between adjacent points. Subsequent innovations like Xu et al.'s (2024) InsMapper explicitly leverage intra-instance query relationships to achieve notable performance gains for HD map detection. Existing architectures either neglect intra-instance spatial dependencies or model them heuristically, leading to suboptimal integration in vectorized systems.

**3. Methodology**

**3.1. Overview**

In this paper, we propose the Intra-instance Vectorized Driving Transformer (InVDriver), a vectorized end-to-end autonomous driving system designed to leverage intra-instance structural relationships across perception, prediction, and planning modules. Different from conventional object detection tasks, these three modules have lines or trajectories as the output, consisting of points and edges. Inspired by VAD (Jiang et al., 2023) and InsMapper (Xu et al., 2024), we use vectors to represent the output of all modules. The data format is visualized in Fig. 1.

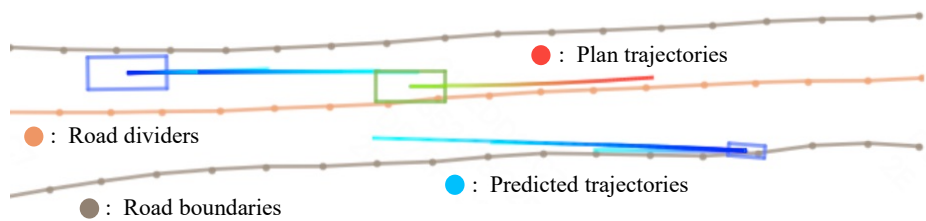

**Fig. 1.** Vectorized representation of end-to-end autonomous driving modules. (1) Perception module. HD map elements (e.g., brown lines show road boundaries and orange lines are road split lines). (2) Prediction module. Blue lines show the predicted trajectories of surrounding objects. (3) Planning module. The red line is the predicted plan trajectory of the ego vehicle. All three modules use vectors to represent the output.

InVDriver comprises three core components: perception, prediction, and planning modules - all employing query-based architectures to generate vectorized outputs that enable seamless integration with downstream modules for end-to-end autonomous driving implementation. The perception module specializes in detecting high-definition (HD) map elements including lane markings, road boundaries, and junction separators. The prediction module produces polyline-formatted forecasts of dynamic agents' future trajectories (e.g., surrounding vehicles), typically generating multiple potential paths per object to account for behavioral uncertainties. The planning module outputs vector-formatted trajectories for the ego vehicle's navigation. All modules employ lightweight vector representations (comprising interconnected points and edges) that preserve critical instance-level geometric information. Current vectorized end-to-end systems typically treat individual points as independent and identically distributed (i.i.d.) elements - an assumption that fundamentally contradicts the inherent spatial correlations between intra-instance points (e.g., consecutive points along a lane marking or vehicle trajectory). To address this limitation, we introduce specialized self-attention mechanisms that explicitly model intra-instance relationships through neighboring point interactions. This architectural enhancement enables coordinated optimization of geometrically coherent points, yielding smoother and more precise linear feature representations. The system architecture is illustrated in Fig. 2.

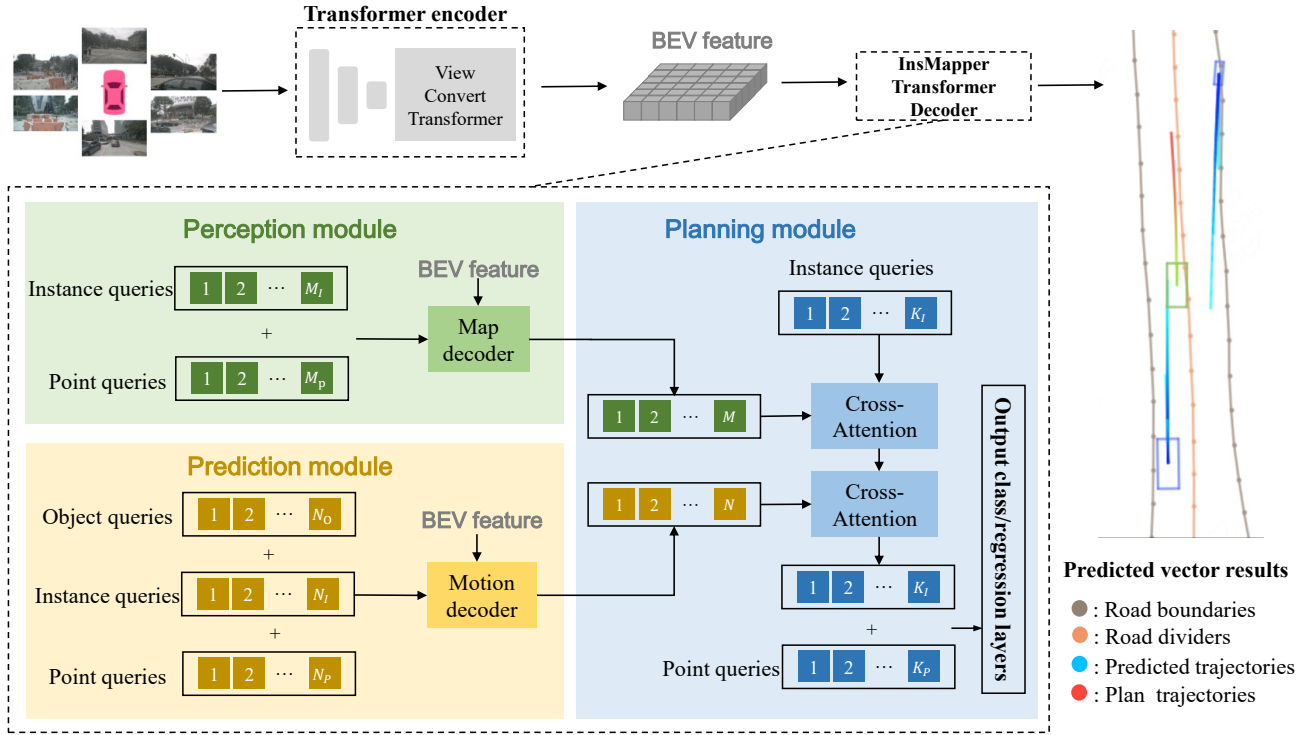

**Fig. 2.** System diagram of InVDriver. InVDriver consists of three modules, i.e., perception (green), prediction (yellow) and planning (blue) modules. All modules are query-based. The perception module extracts vectorized HD map elements through structured query processing, while the prediction module generates multi-modal trajectory forecasts for surrounding agents. The planning module performs cross-attention operations on these processed environmental representations, synthesizing multi-source inputs to generate optimized ego-vehicle trajectories.

## 3.2 System design

In this section, we introduce the design of our systems with details. Designed for efficient multi-sensor processing, our system ensures seamless information flow across perception, prediction, and planning. By leveraging vectorized representations and cross-attention mechanisms, it enhances spatial consistency and temporal coherence, addressing key challenges in end-to-end autonomous driving.

### 3.2.1. BEV encoder

The system processes multi-camera inputs through a transformer encoder to generate bird's-eye-view (BEV) feature representations. The resultant BEV tensor encodes spatial information of the ego vehicle's surroundings in rasterized format, serving three critical functions: 1) providing geometric priors for perception module queries, 2) enabling temporal reasoning in prediction module operations, and 3) facilitating cross-modal attention mechanisms between perception-prediction subsystems. This structured representation forms the computational substrate for subsequent cross-attention operations across perception and prediction modules.

### 3.2.2. Perception module

The perception module specializes in vectorized HD map detection, operating through two distinct query types: instance queries and point queries. The former encodes individual HD map elements at the instance level (e.g., complete lane markings or road boundaries), while the latter parameterizes geometric details of constituent points within each instance. This dual-query architecture enables hierarchical representation learning that preserves both instance semantics and point-level geometric precision. Suppose there are $M_I$ instance queries and $M_P$ point queries, and the input queries of the map decoder is the pairwise addition of two kinds of queries, then there will be $M_I \times M_P$ input queries. To better utilize intra-instance information, we add additional masked self-attention modules during query initialization and after the cross-attention layer of the decoder. The mask of the self-attention module blocks information exchange between inter-instance points (i.e., points of different instance) and only allows information exchange between intra-instance points (i.e., points of the same instance). The refined transformer network structure is visualized in Fig. 3. The masked self-attention mechanism is shown in Fig. 4.

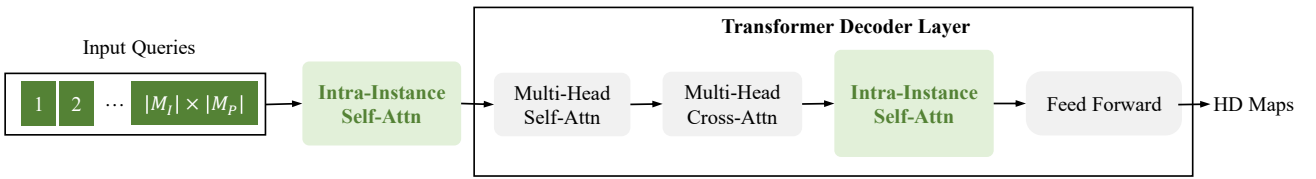

**Fig. 3.** The query-based perception module. It has been refined for better utilizing the intra-instance information. The intra-instance query initialization module and intra-instance self-attention module are highlighted in green color. Both modules are realized by masked self-attention, where the mask blocks inter-instance information and only allows intra-instance information exchange.

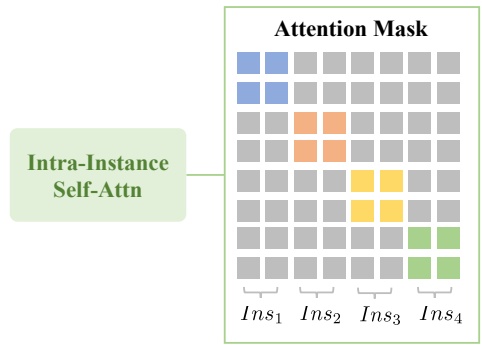

**Fig. 4.** The intra-instance self-attention layer. This module is masked by specially defined mask, which blocks information exchange between inter-instance points (i.e., points of different instance, grey grids in the mask) and only allows information exchange between intra-instance points (i.e., points within the same instance, colored grids in the mask).

### 3.2.3. Prediction module

The prediction module is responsible for detecting surrounding dynamic agents, such as vehicles and pedestrians, and forecasting their potential motion trajectories. Accurate motion prediction is critical for ensuring safe and efficient decision-making in autonomous driving, particularly in highly interactive traffic environments. Suppose the number of detected objects as $N_O$. Each trajectory is represented by a fixed-length polyline with $N_P$ points. To show all possible motions, we predict $N_I$ trajectories for each predicted object. Thus there are finally $N_O \times N_I \times N_P$ input queries for the motion decoder, and each query represents a single point. To ensure spatial and temporal coherence in predicted trajectories, the prediction module employs a masked self-attention mechanism, similar to the perception module. By restricting attention to intra-instance interactions, the model effectively refines trajectory predictions, mitigating noise and improving robustness in highly dynamic environments.

### 3.2.4. Planning module

The planning module's decision-making process demonstrates critical dependence on both HD map topology and dynamic agent trajectories. This architectural design consequently implements cross-modal attention mechanisms between the perception module's vectorized map features and the prediction module's motion forecasting queries, strategically bypassing conventional BEV tensor processing. Through direct query-level interaction between environmental understanding and behavioral prediction subsystems, the system achieves context-aware trajectory generation while maintaining computational efficiency through structured information fusion. Suppose there are $K_I$ possible motion trajectories of the ego vehicle, and each trajectory contains $K_P$ points, then there will be $K_I \times K_P$ input queries. They are processed similar to that of the perception module to better utilize the intra-instance information.

### 3.3. Loss functions

The proposed InVDriver framework is trained in an end-to-end manner, leveraging a combination of multiple loss functions to ensure both geometric consistency and trajectory accuracy. The optimization process is governed by three primary L1-distance losses, which enforce point-wise accuracy across different modules, and a vectorized constraint loss, as introduced in VAD (Jiang et al., 2023), to preserve the structural coherence of vectorized outputs. The overall loss function integrates the above components in a weighted sum.

By combining point-wise accuracy (L1-loss) with structural regularization (vectorized constraint loss), our approach ensures that InVDriver produces highly precise and geometrically consistent outputs, leading to more stable and interpretable autonomous driving decisions.

## 4. Experiments and discussions

### 4.1. Dataset

Our experimental validation leverages the nuScenes benchmark (Caesar et al., 2020), a widely-recognized autonomous driving dataset comprising 1,000 diverse 20-second driving sequences. This benchmark provides comprehensive scene understanding data with 1.4 million annotated 3D bounding boxes across 23 object categories. The visual perception foundation is established through 360° horizontal coverage from six synchronized cameras, with critical scene elements annotated at 2Hz temporal resolution. For planning performance quantification, we employ two principal metrics: Displacement Error (DE) measuring trajectory deviation accuracy, and Collision Rate (CR) evaluating obstacle avoidance capability. At the same time, we report the detection performance of the HD map detection task, demonstrating the powerful overall performance of the proposed system.

### 4.2. Experiment settings

We perform all experiments on 8 V100 GPUs, and test the model on a single A800 GPU. For the perception module, we set the number of instance values and points values as 100 and 20, which means there are 2,000 input queries for the map decoder. For the prediction module, each object predicts 5 possible trajectories, and each trajectory contains 6 points. Finally, the planning module has 3 instance queries to predict 3 possible trajectories, and each trajectory have 6 points. The end-to-end learning rate is set to $2^{-4}$, and the model is trained for 60 epochs.

**Table 1** Evaluation results on the NuScenes dataset. Displacement Error (DE) is evaluated by L2 distance (m), measuring the different between the predicted and ground-truth trajectories. Collision Rate (CR) is evaluated by collision rate (%). Lower number represents better performance for both DE and CR. FPS shows model efficiency. Grey rows are methods with both and LiDAR and cameras, while other rows show methods only with cameras.

| Method | L2(m) ↓ | | | | Collision(%) ↓ | | | | FPS |
|---|---|---|---|---|---|---|---|---|---|
| | 1s | 2s | 3s | Avg. | 1s | 2s | 3s | Avg. | |
| NMP (Zeng et al., 2019) | - | - | 2.31 | - | - | - | 1.92 | - | - |
| SA-NMP (Zeng et al., 2019) | - | - | 2.05 | - | - | - | 1.59 | - | - |
| FF (Hu et al., 2021) | 0.55 | 1.20 | 2.54 | 1.43 | 0.06 | 0.17 | 1.07 | 0.43 | - |

| EO (Khurana et al., 2022) | 0.67 | 1.36 | 2.78 | 1.60 | 0.04 | 0.09 | 0.88 | 0.33 | - |
| ST-P3 (Hu et al., 2022) | 1.33 | 2.11 | 2.90 | 2.11 | 0.23 | 0.62 | 1.27 | 0.71 | 1.6 |
| UniAD (Hu et al., 2023) | 0.48 | 0.96 | 1.65 | 1.03 | **0.05** | **0.17** | 0.71 | 0.31 | 1.8 |
| VAD (Jiang et al., 2023) | 0.46 | 0.76 | 1.12 | 0.78 | 0.21 | 0.35 | 0.58 | 0.38 | **16.8** |
| InVDriver | **0.26** | **0.46** | **0.78** | **0.50** | 0.14 | 0.23 | **0.48** | **0.28** | 15.3 |

### 4.3. Comparison results

Our proposed InVDriver is rigorously evaluated on the nuScenes benchmark against state-of-the-art planning methods, including LiDAR-camera fusion and camera-only approaches. As shown in Table 1, InVDriver achieves superior trajectory accuracy with displacement errors (DE) of 0.26m/0.46m/0.78m at 1s/2s/3s horizons, outperforming the previous best vision-based method VAD by 36% in average DE. This improvement stems from the novel intra-instance attention mechanism, which effectively coordinates geometrically coherent points within trajectories, addressing the limitations of independent point modeling in conventional vectorized systems. Notably, InVDriver also sets new safety benchmarks, reducing collision rates (CR) to 0.14%/0.23%/0.48% across time horizons—a 26% average CR reduction compared to VAD. This breakthrough underscores the efficacy of cross-module attention in explicitly encoding HD map constraints and dynamic agent interactions, thereby mitigating risky maneuvers in complex scenarios like unprotected turns and pedestrian-dense zones. Despite these advancements, InVDriver maintains high efficiency at 15.3 FPS, comparable to other camera-only methods (e.g., UniAD: 1.8 FPS, VAD: 16.8 FPS). Such efficiency is attributed to our lightweight query propagation design, which can effectively boost the planning results and maintain the competitive efficency. Further analysis reveals two critical insights: First, vision-only methods consistently surpass LiDAR-fusion systems in both accuracy and safety, validating the superiority of vectorized representations in compressing geometric priors. Second, InVDriver's widening performance gap at longer horizons (e.g., 3s DE: 0.78m vs. VAD's 1.12m) highlights its exceptional temporal consistency, a direct result of structured intra-instance optimization. These results collectively demonstrate that InVDriver successfully balances safety-critical performance, trajectory precision, and computational efficiency, offering a practical framework for deployable autonomous driving systems.

### 4.4 Ablation studies

**Table 2** Ablation studies. We evaluate the proposed intra-instance (i.e., intra-ins.) modules for better intra-instance information utilization, as well as the mask of added self-attention modules (i.e., use original self-attention to replace masked self-attention). "√" means the design is applied.

| Perception intra-ins. | Prediction intra-ins | Planning intra-ins | Mask self-attn. | L2(m) ↓ | | | | Collision(%) ↓ | | | |
|---|---|---|---|---|---|---|---|---|---|---|---|
| | | | | 1s | 2s | 3s | Avg. | 1s | 2s | 3s | Avg. |
| | | | | 0.46 | 0.76 | 1.12 | 0.78 | 0.21 | 0.35 | 0.58 | 0.38 |
| √ | √ | | √ | 0.45 | 0.63 | 1.06 | 0.71 | 0.23 | 0.31 | 0.55 | 0.36 |
| | √ | √ | √ | 0.30 | 0.54 | 0.89 | 0.58 | 0.20 | 0.29 | 0.51 | 0.33 |
| √ | | √ | √ | 0.29 | 0.49 | 0.82 | 0.53 | **0.11** | 0.26 | 0.56 | 0.31 |
| √ | √ | √ | | 0.49 | 0.72 | 1.18 | 0.80 | 0.23 | 0.41 | 0.65 | 0.43 |
| √ | √ | √ | √ | **0.26** | **0.46** | **0.78** | **0.50** | 0.14 | **0.23** | **0.48** | **0.28** |

To validate the effectiveness of our proposed components, we conduct systematic ablation experiments analyzing the contributions of intra-instance optimization and masked self-attention mechanisms across perception, prediction, and planning modules. As shown in Table 2, removing intra-instance modules from any autonomous driving modules will notably degrade the final performance, including both DE and CR. Therefore, the intra-instance designs added into all three modules are justified and cannot be removed. We also evaluate the necessity of the masked self-attention. The intra-instance modules is realized by masked self-attention, which can block inter-instance information exchange and only allow intra-instance information exchange. The masked self-attention is critical for better intra-instance information utilization. If we use normal self-attention to replace the masked self-attention, all intra-instance modules would not function properly. From the results, we can see that removing the masked self-attention design will greatly harm the final performance. These experiments conclusively prove that our intra-instance attention paradigm fundamentally enhances vectorized autonomy systems by enabling structured point interactions while maintaining computational efficiency through context-aware attention masking. Therefore, the design of InVDriver is justified.

**4.5 Qualitative results**

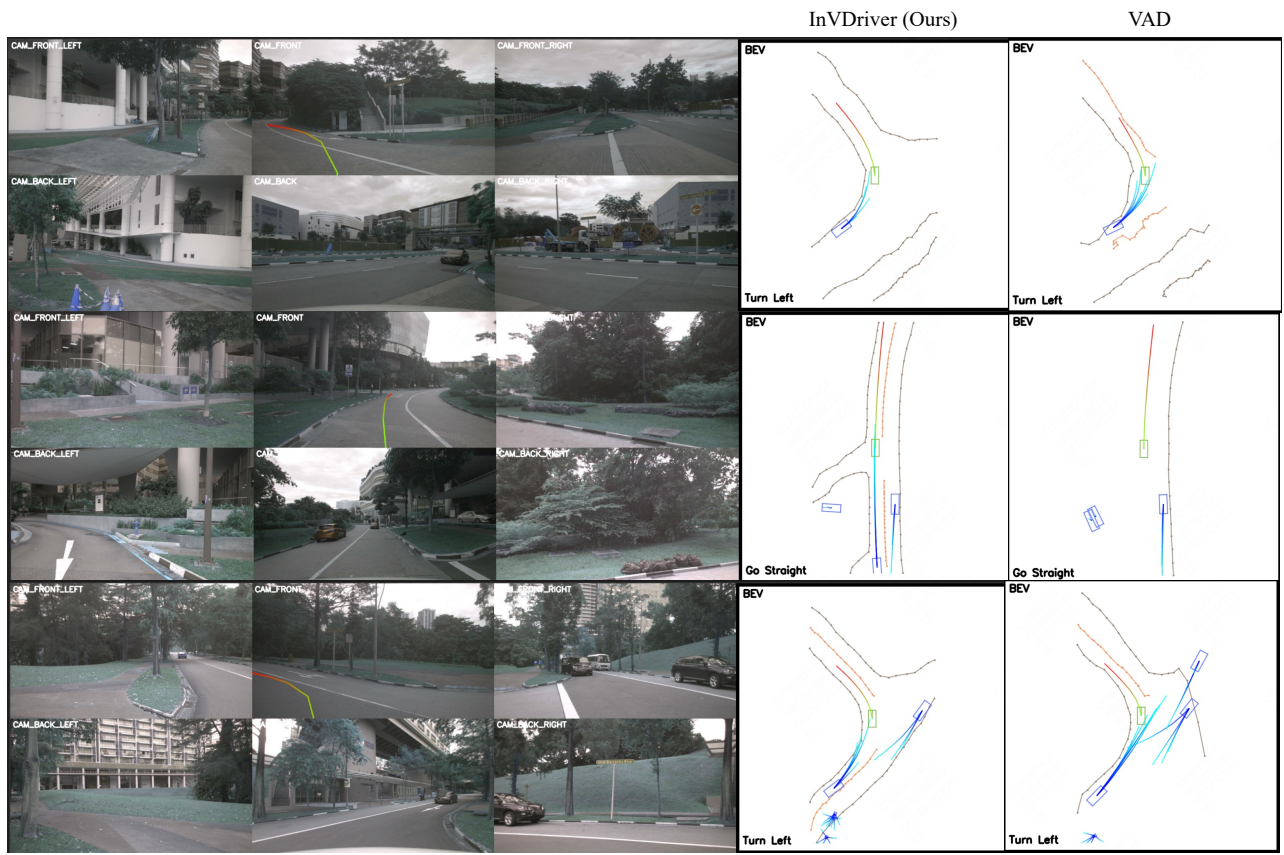

**Fig. 5.** Qualitative comparison of InVDriver and VAD in diverse scenarios.

We provide multiple qualitative visualization results to better compare our proposed InVDriver and previous SOTA method VAD. With multiple camera images as input, both methods output HD map predictions, surrounding vehicle motions and the trajectory of ego vehicle. From Fig. 5, InVDriver produces smoother and more geometrically consistent HD map predictions, effectively capturing lane boundaries and road structures with high precision. In contrast, VAD exhibits zig-zag distortions in its HD map outputs and even fails to detect certain road elements, which could compromise downstream planning.

Additionally, the motion prediction results of surrounding vehicles highlight a significant difference in output quality. While InVDriver generates structured and stable trajectories, VAD's predictions appear noisy and inconsistent, potentially leading to suboptimal planning decisions. By leveraging intra-instance relationships and improved self-attention mechanisms, InVDriver ensures more reliable perception and prediction outputs, ultimately leading to safer and more efficient motion planning.

## 5. Conclusion and future work

In this work, we present InVDriver, a novel vectorized query-based end-to-end autonomous driving framework comprising three interconnected modules—perception, prediction, and planning—unified through learnable queries and vectorized representations. The core innovation lies in our intra-instance masked self-attention mechanism, which selectively restricts information exchange to geometrically coherent points within structural elements (e.g., HD map lanes, agent trajectories) while suppressing irrelevant inter-instance interactions. This design enables systematic refinement of linear features, yielding smoother and more precise outputs compared to conventional vectorized approaches. Extensive evaluations on the nuScenes benchmark demonstrate state-of-the-art performance while maintaining competitive efficiency through lightweight architectural design. Despite these advancements, our analysis reveals limitations in modeling complex scene topology, where ambiguous lane connectivity or dynamic occlusion patterns may degrade planning reliability. Future work will integrate topology-aware graph neural networks to model lane connectivity constraints and explore real-time deployment feasibility under computational constraints.

**Acknowledgements**

This research was supported by National Natural Science Foundation of China, Science Fund for Creative Research Groups (52221005) and National Natural Science Foundation of China, the Key Project (52131201).

**Declaration of competing interest**

We declare that we have no known competing financial interests or personal relationships that could have appeared to influence the work reported in this paper.

**Reference**

Jiang, B., Chen, S., Xu, Q., Liao, B., Chen, J., Zhou, H., Zhang, Q., Liu, W., Huang, C. and Wang, X., 2023. Vad: Vectorized scene representation for efficient autonomous driving. In Proceedings of the IEEE/CVF International Conference on Computer Vision (pp. 8340-8350).

Pan, C., Yaman, B., Nesti, T., Mallik, A., Allievi, A.G., Velipasalar, S. and Ren, L., 2024. VLP: Vision Language Planning for Autonomous Driving. In Proceedings of the IEEE/CVF Conference on Computer Vision and Pattern Recognition (pp. 14760-14769).

Xiao, Y., Codevilla, F., Gurram, A., Urfalioglu, O. and López, A.M., 2020. Multimodal end-to-end autonomous driving. IEEE Transactions on Intelligent Transportation Systems, 23(1), pp.537-547.

Zheng, W., Song, R., Guo, X., Zhang, C. and Chen, L., 2024, September. Genad: Generative end-to-end autonomous driving. In European Conference on Computer Vision (pp. 87-104). Cham: Springer Nature Switzerland.

Hu, Y., Yang, J., Chen, L., Li, K., Sima, C., Zhu, X., Chai, S., Du, S., Lin, T., Wang, W. and Lu, L., 2023. Planning-oriented autonomous driving. In Proceedings of the IEEE/CVF Conference on Computer Vision and Pattern Recognition (pp. 17853-17862).

Zeng, W., Luo, W., Suo, S., Sadat, A., Yang, B., Casas, S. and Urtasun, R., 2019. End-to-end interpretable neural motion planner. In Proceedings of the IEEE/CVF Conference on Computer Vision and Pattern Recognition (pp. 8660-8669).

Hu, P., Huang, A., Dolan, J., Held, D. and Ramanan, D., 2021. Safe local motion planning with self-supervised freespace forecasting. In Proceedings of the IEEE/CVF Conference on Computer Vision and Pattern Recognition (pp. 12732-12741).

Hu, S., Chen, L., Wu, P., Li, H., Yan, J. and Tao, D., 2022, October. St-p3: End-to-end vision-based autonomous driving via spatial-temporal feature learning. In European Conference on Computer Vision (pp. 533-549). Cham: Springer Nature Switzerland.

Khurana, T., Hu, P., Dave, A., Ziglar, J., Held, D. and Ramanan, D., 2022, October. Differentiable raycasting for self-supervised occupancy forecasting. In European Conference on Computer Vision (pp. 353-369). Cham: Springer Nature Switzerland.

Liao, B., Chen, S., Wang, X., Cheng, T., Zhang, Q., Liu, W. and Huang, C., 2022. Maptr: Structured modeling and learning for online vectorized hd map construction. arXiv preprint arXiv:2208.14437.

Xu, Z., K. Wong, K.Y. and Zhao, H., 2024, September. InsMapper: Exploring inner-instance information for vectorized HD mapping. In European Conference on Computer Vision (pp. 296-312). Cham: Springer Nature Switzerland.

Xu, Z., Liu, Y., Gan, L., Sun, Y., Wu, X., Liu, M. and Wang, L., 2022. Rngdet: Road network graph detection by transformer in aerial images. IEEE Transactions on Geoscience and Remote Sensing, 60, pp.1-12.

Xu, Z., Liu, Y., Sun, Y., Liu, M. and Wang, L., 2023. Rngdet++: Road network graph detection by transformer with instance segmentation and multi-scale features enhancement. IEEE Robotics and Automation Letters, 8(5), pp.2991-2998.

Xu, Z., Zhang, Y., Xie, E., Zhao, Z., Guo, Y., Wong, K.Y.K., Li, Z. and Zhao, H., 2024. Drivegpt4: Interpretable end-to-end autonomous driving via large language model. IEEE Robotics and Automation Letters.

Xu, Z., Sun, Y. and Liu, M., 2021. icurb: Imitation learning-based detection of road curbs using aerial images for autonomous driving. IEEE Robotics and Automation Letters, 6(2), pp.1097-1104.

Xu, Z., Sun, Y. and Liu, M., 2021. Topo-boundary: A benchmark dataset on topological road-boundary detection using aerial images for autonomous driving. IEEE Robotics and Automation Letters, 6(4), pp.7248-7255.

Chen, J., Xiang, Y., Luo, Y., Li, K. and Lian, X., 2024. Decision making and control of autonomous vehicles under the condition of front vehicle sideslip. Journal of Intelligent and Connected Vehicles, 7(4), pp.248-257.

Zhang, Y., Tu, C., Gao, K. and Wang, L., 2024. Multisensor information fusion: Future of environmental perception in intelligent vehicles. Journal of Intelligent and Connected Vehicles.

Berta, R., Lazzaroni, L., Capello, A., Cossu, M., Forneris, L., Pighetti, A. and Bellotti, F., 2024. Development of deep-learning-based autonomous agents for low-speed maneuvering in Unity. Journal of Intelligent and Connected Vehicles, 7(3), pp.229-244.

Li, Q., Zhang, P., Yao, H., Chen, Z. and Li, X., 2024. Online learning-based model predictive trajectory control for connected and autonomous vehicles: Modeling and physical tests. Journal of Intelligent and Connected Vehicles, 7(2), pp.86-96.

Zheng, X., Li, H., Qiang, Z., Liu, Y., Chen, X., Liu, H., Luo, T., Gao, J. and Xia, L., 2024. Intelligent decision-making method for vehicles in emergency conditions based on artificial potential fields and finite state machines. Journal of Intelligent and Connected Vehicles.

Singh, A., Yahoodik, S., Murzello, Y., Petkac, S., Yamani, Y. and Samuel, S., 2024. Ethical decision-making in older drivers during critical driving situations: An online experiment. Journal of Intelligent and Connected Vehicles.

Wang, Y., Ho, I.W.H. and Wang, Y., 2023. Real-time intersection vehicle turning movement counts from live UAV video stream using multiple object tracking. Journal of Intelligent and Connected Vehicles.

Cao, L., Luo, Y., Wang, Y., Chen, J. and He, Y., 2023. Vehicle sideslip trajectory prediction based on time-series analysis and multi-physical model fusion. Journal of Intelligent and Connected Vehicles.

Shao, H., Hu, Y., Wang, L., Song, G., Waslander, S.L., Liu, Y. and Li, H., 2024. Lmdrive: Closed-loop end-to-end driving with large language models. In Proceedings of the IEEE/CVF Conference on Computer Vision and Pattern Recognition (pp. 15120-15130).

Sima, C., Renz, K., Chitta, K., Chen, L., Zhang, H., Xie, C., Beißwenger, J., Luo, P., Geiger, A. and Li, H., 2024, September. Drivelm: Driving with graph visual question answering. In European Conference on Computer Vision (pp. 256-274). Cham: Springer Nature Switzerland.

Liu, T., hai Liao, Q., Gan, L., Ma, F., Cheng, J., Xie, X., Wang, Z., Chen, Y., Zhu, Y., Zhang, S. and Chen, Z., 2021. The role of the hercules autonomous vehicle during the covid-19 pandemic: An autonomous logistic vehicle for contactless goods transportation. IEEE Robotics & Automation Magazine, 28(1), pp.48-58.

Liu, Y., Xu, Z., Huang, H., Wang, L. and Liu, M., 2023. FSNet: Redesign Self-Supervised MonoDepth for Full-Scale Depth Prediction for Autonomous Driving. IEEE Transactions on Automation Science and Engineering.

Zablocki, É., Ben-Younes, H., Pérez, P. and Cord, M., 2022. Explainability of deep vision-based autonomous driving systems: Review and challenges. International Journal of Computer Vision, 130(10), pp.2425-2452.

Kendall, A., Hawke, J., Janz, D., Mazur, P., Reda, D., Allen, J.M., Lam, V.D., Bewley, A. and Shah, A., 2019, May. Learning to drive in a day. In 2019 international conference on robotics and automation (ICRA) (pp. 8248-8254). IEEE.

Li, Q., Wang, Y., Wang, Y. and Zhao, H., 2022, May. Hdmapnet: An online hd map construction and evaluation framework. In 2022 International Conference on Robotics and Automation (ICRA) (pp. 4628-4634). IEEE.

Liu, Y., Yuan, T., Wang, Y., Wang, Y. and Zhao, H., 2023, July. Vectormapnet: End-to-end vectorized hd map learning. In International Conference on Machine Learning (pp. 22352-22369). PMLR.

Xu, Z., Liu, Y., Sun, Y., Liu, M. and Wang, L., 2023, May. Centerlinedet: Centerline graph detection for road lanes with vehicle-mounted sensors by transformer for hd map generation. In 2023 IEEE International Conference on Robotics and Automation (ICRA) (pp. 3553-3559). IEEE.

Caesar, H., Bankiti, V., Lang, A.H., Vora, S., Liong, V.E., Xu, Q., Krishnan, A., Pan, Y., Baldan, G. and Beijbom, O., 2020. nuscenes: A multimodal dataset for autonomous driving. In Proceedings of the IEEE/CVF conference on computer vision and pattern recognition (pp. 11621-11631).

**Author biography**

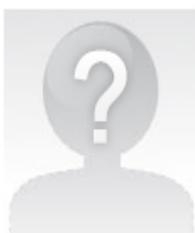

**Author name**  Biography shows each author's biography, including the photo, educational background, working experience, research field, etc.